\documentclass{article}

\PassOptionsToPackage{numbers}{natbib} % For numerical references
\usepackage[preprint]{nips_2018}

\usepackage[utf8]{inputenc} % allow utf-8 input
\usepackage[T1]{fontenc}    % use 8-bit T1 fonts
\usepackage{hyperref}       % hyperlinks
\usepackage{url}            % simple URL typesetting
\usepackage{booktabs}       % professional-quality tables
\usepackage{amsfonts}       % blackboard math symbols
\usepackage{nicefrac}       % compact symbols for 1/2, etc.
\usepackage{microtype}      % microtypography

\usepackage{graphicx} % more modern
\usepackage{amsmath}
\usepackage{caption}
\usepackage{subcaption}

\usepackage{algorithm}
\usepackage{algorithmic}
\usepackage{amssymb}
\newcommand{\argmin}[1]{\underset{#1}{\operatorname{argmin}}\;}
\newcommand{\argmax}[1]{\underset{#1}{\operatorname{argmax}}\;}
\def\spose#1{\hbox to 0pt{#1\hss}}
\def\simlt{\mathrel{\spose{\lower 3pt\hbox{$\mathchar"218$}}
   \raise 2.0pt\hbox{$\mathchar"13C$}}}
\def\simgt{\mathrel{\spose{\lower 3pt\hbox{$\mathchar"218$}}
     \raise 2.0pt\hbox{$\mathchar"13E$}}}
 \def\simpropto{\mathrel{\spose{\lower 3pt\hbox{$\mathchar"218$}}
     \raise 2.0pt\hbox{$\propto$}}}

\def\beq#1{\begin{equation}\label{#1}}
\def\eeq{\end{equation}}
\def\beqa#1{\begin{eqnarray}\label{#1}}
\def\eeqa{\end{eqnarray}}

\def\fig#1{Figure~\ref{#1}}

\def\D{{\bf D}}
\def\f{{\bf f}}
\def\g{{\bf g}}
\def\h{{\bf h}}
\def\J{{\bf J}}
\def\m{{\bf m}}
\def\x{{\bf x}}
\def\y{{\bf y}}
\def\yhat{\hat{\bf y}}
\def\z{{\bf z}}
\def\X{{\bf X}}
\def\Y{{\bf Y}}

\def\gammavec{\boldsymbol\gamma}
\def\phivec{\boldsymbol\theta}
\def\muvec{\boldsymbol\mu}

\newcommand{\R}{\mathbb{R}}
\def\sout{s_{\rm out}}
\def\spool{s_{\rm pool}}
\def\slat{s_{\rm code}}

\def\ie{{\frenchspacing\it i.e.}}
\def\eg{{\frenchspacing\it e.g.}}

\title{Meta-learning autoencoders for few-shot prediction}

\author{
  Tailin Wu \\
  MIT\\
  \texttt{tailin@mit.edu} \\
  \And
  John Peurifoy\\
  MIT\\
  \texttt{jpeurifo@mit.edu}
   \AND
  Isaac L. Chuang\\
  MIT\\
  \texttt{ichuang@mit.edu}
   \And
  Max Tegmark\\
  MIT\\
  \texttt{tegmark@mit.edu}
}

\begin{document}

\maketitle

\begin{abstract}

Compared to humans, machine learning models generally require significantly more training examples and fail to extrapolate from experience to solve previously unseen challenges.  To help close this performance gap, we augment single-task neural networks with a meta-recognition model which learns a succinct model code via its autoencoder structure, using just a few informative examples. The model code is then employed by a meta-generative model to construct parameters for the task-specific model. We demonstrate that for previously unseen tasks, without additional training, this {\it Meta-Learning Autoencoder} (MeLA) framework can build models that closely match the true underlying models, with loss significantly lower than given by fine-tuned baseline networks, and performance that compares favorably with state-of-the-art meta-learning algorithms.  MeLA also adds the ability to identify influential training examples and 
%by interactively request  optimum input for optimal learning.
predict which additional data will be most valuable to acquire to improve model prediction.

% After merely a quick glance at a new environment, humans can propose a reasonable model based on previously learned models. In contrast, machine learning models typically require a large number of examples to train, and cannot readily use transfer past knowledge to unseen tasks. We propose \emph{Meta-Learning Autoencoders} (MeLA), a framework that can transform a model originally intended for single-task learning into one that can quickly (without training) adapt to new tasks with few examples. MeLA consists of a meta-recognition model that calculates a model code based on a few informative examples, and a meta-generative model that generates the parameters of a task-specific neural network based on the model code. We demonstrate that for new unseen tasks, MeLA can without training propose models significantly more accurate than a fine-tuned baseline network, significantly faster than the state-of-the-art meta-learning algorithm MANL. We also demonstrate MeLA's ability to preduct which additional data will be most valuable to acqquire to improve model prediction.
\end{abstract}

% \subsection{Things to think about or improve}
% Fig 3(a) legends
% Add maml for fig 5
% Add RL experiments
% Reference: Meta networks

% Can you find a bib style whose citation is like [1,2]? 

% (1) How to justify that we make the assumption that the hidden parameter vector has not too large dimensions? 
    
%   (2) How large is large?
   
%   (3) Is there a way from the model to tell whether the dimension you use for the latent parameter is large enough? JUST TRY DIFFERENT SIZES n \& PLOT LOSS VS n
   
%   I think another way is to perform PCA on the latent code for different tasks. 
%   But it could be that a nonlinear autoencoder could compress the latent variables further. PCA only discovers linear compression opportunities.
  
% 3. Add some succinct comparison with other models in introduction.

% Things to fix:
% 1. True model thicker dashed.
% 1. HyperNEAT reference
% 2. Consistency of symbols in figures
% 4. Font in the figure
% 6. Add maml in the figures.
% 7. Abstract

\section{Introduction}
A key ingredient of human intelligence is the ability to generalize beyond models that have already been learned, and quickly propose new models in new environments with few examples. For example, after seeing several moving objects accelerated by different forces in different environments,
% Just talking about acceleration is progblematic and likely to confuse readers, since most humans alsways see the same g.
humans are able to not only develop models for each environment, but also to develop a meta-model that can generalize to a continuum of unseen accelerated objects. Upon arriving at a new environment and seeing an object moving for only a short time, s/he can quickly propose a new model for the moving object, including a good estimate of its acceleration. Incorporating this ability of generalization beyond the training datasets and environments for quick recognition from few examples remains an important challenge in machine learning.

In this paper, we focus on learning a series of prediction/regression models with continuous targets, where each class of problems has similar underlying mechanisms. Algorithms are compared by how well and how quickly they can generalize to unseen tasks from few examples. This class of problems is important in many areas, for example, learning and predicting physics \cite{NIPS2017_6951,NIPS2017_6620,NIPS2017_7040,chang2016compositional}, reinforcement learning of games \cite{mnih2015human,Kansky2017SchemaNZ}, unsupervised learning of videos \cite{Srivastava:2015:ULV:3045118.3045209}, and applications such as self-driving, where we cannot enumerate and train with all environments that the algorithm will encounter. 

To tackle such problems, we propose a novel class of neural networks that we term {\it Meta-Learning Autoencoders} (MeLA), schematically
 illustrated in \fig{ArchitectureFig}.
At its core, a MeLA consists of a learnable meta-recognition model that can for each (unseen) task distill a few input-output examples into a model code vector parametrizing the task's functional relationship, and a learnable meta-generative model that maps this model code into the weight and bias parameters of a neural network implementing this function. This architecture forces the meta-recognition model to discover and encode the important variations of the functional mappings for different tasks, and the meta-generative model to decode the model codes to corresponding task-specific models with a common model-generating network. This brings the key innovation of MeLA: for a class of tasks, MeLA does not attempt to learn a \emph{single} good initialization for multiple tasks as in \cite{pmlr-v70-finn17a}, or learn an update function \cite{schmidhuber1987evolutionary,bengio1992optimization,andrychowicz2016learning}, or learn an update function together with a \emph{single} good initialization \cite{li2017meta, ravi2016optimization}. Instead, it learns to map the few examples from different datasets into \emph{different} models, which not only allows for more diverse model parameters tailored for each individual tasks, but also obviates the need for fine-tuning.
% This endows to the key innovation of MeLA: MeLA is able to propose models that unlike prior methods of meta-learning that learn a single ``sweet spot" of model parameter that can easily fine-tune to individual tasks , or learn a learning rule or update function \cite{}, or learn an update function from a single sweet spot \cite{}, MeLA to propose models that may have % A key innovation of this architecture is that it learns to directly ``jump" to vastly different model parameters based on a few examples, 
% Also, in contrast to the neural statistician approach \cite{edwards2016towards} that uses the VAE technique to generate the \emph{dataset}, we use an autoencoding architecture to generate the \emph{model} that can generate the dataset given test inputs, since in general we don't know \emph{a priori} where the input will be in testing time. 
Moreover, by encoding each function as a vector in a single low-dimensional latent space, MeLA is able to generalize beyond the training datasets, by both interpolating between and extrapolating beyond learned models into a continuum of models. 
% We propose two training methods: joint training and variational training. 
%In the latter case, it becomes a variational meta-learning autoencoder. 
We will demonstrate that the meta-learning autoencoder has the following 3 important capabilities:
\begin{enumerate}
    \item \textbf{Augmented model recognition:} MeLA strategically builds on a pre-existing, single-task trained network, augmenting it with a second network used for meta-recognition. It achieves lower loss in unseen environments at zero and few gradient steps, compared with both the original architecture upon which it is based and state-of-the-art meta-learning algorithms.
   % and fine-tuning baselines
    % \item propose a reasonable model with a few examples even in unseen environments whose mechanism lies beyond the scope of training datasets, where other algorithms fail,       
    \item \textbf{Influence identification:} MeLA can identify which examples are most useful for determining the model
  (for example, a rectangle's vertices have far greater influence in determining its position and size than its inferior points).
  
    \item \textbf{Interactive learning:} MeLA can actively request new samples which maximize its ability to learn models.
\end{enumerate}
\section{Methods}
\label{sec:Methods}

%\subsection{Meta-learning problem setup}
\paragraph{Meta-learning problem setup}

We are interested in modeling a set of vector-valued functions $\h_{\alpha}$ (which we will refer to as {\it models}), that each map an $m$-dimensional input vector $\x$ into an $n$-dimensional output vector $\y$.
Let's first consider the case for a single dataset. Given many input-output pairs 
$\y_i=\h_\alpha(\x_i)$ linked by the same function $\h_{\alpha}$, we group the corresponding vectors into matrices 
$\X$ and $\Y$ whose $i^{th}$ rows are the vectors $\x_i$ and $\y_i$. In this paper, we focus on regression problems where the target $\y\in\R^n$ is continuous, but the generalization to classification problems is straightforward. This class of problems includes a wide range of scenarios, {\eg}, modeling time series data, learning physics and dynamics, and frame-to-frame prediction of videos.

The meta-learning problem we tackle is as follows. Suppose that we are given an ensemble of datasets 
$\{\D^\alpha\}=\{(\X^\alpha,\Y^\alpha)\}$, 
$\alpha=1,2,...$, each of which is generated by a corresponding function $\h_\alpha$. 
In the single-task scenario, we want to train a model $\f$ 
that predicts all output vectors from the corresponding input vectors from a single dataset, 
to minimize some loss function $\ell$ that quantifies the prediction errors. 
%$\Phi$ is the hidden parameter vector of the hidden model $g_\Phi$. 
The meta-learning goal is, after training on an ensemble of training datasets 
$\D^1$,...,$\D^a$,
%$\{\D^\alpha\}$, $\alpha=1,...,a$, 
to be able to quickly learn from few examples from held-out datasets 
$\D^{a+1}$,...,
%$\{\D^\alpha\}$, $\alpha=a+1,...$,
adapt to them and obtain a low loss on them.

% In this paper, we further assume that the dimension of $\Phi$ is not too large. In other words, the ensemble of datasets have some similar underlying mechanisms. 
% Due to the no free lunch theorem, 

\begin{figure}[t]
\begin{center}
% \fbox{\rule[-.5cm]{0cm}{4cm} \rule[-.5cm]{4cm}{0cm}}
\centerline{\includegraphics[width=0.8\columnwidth]{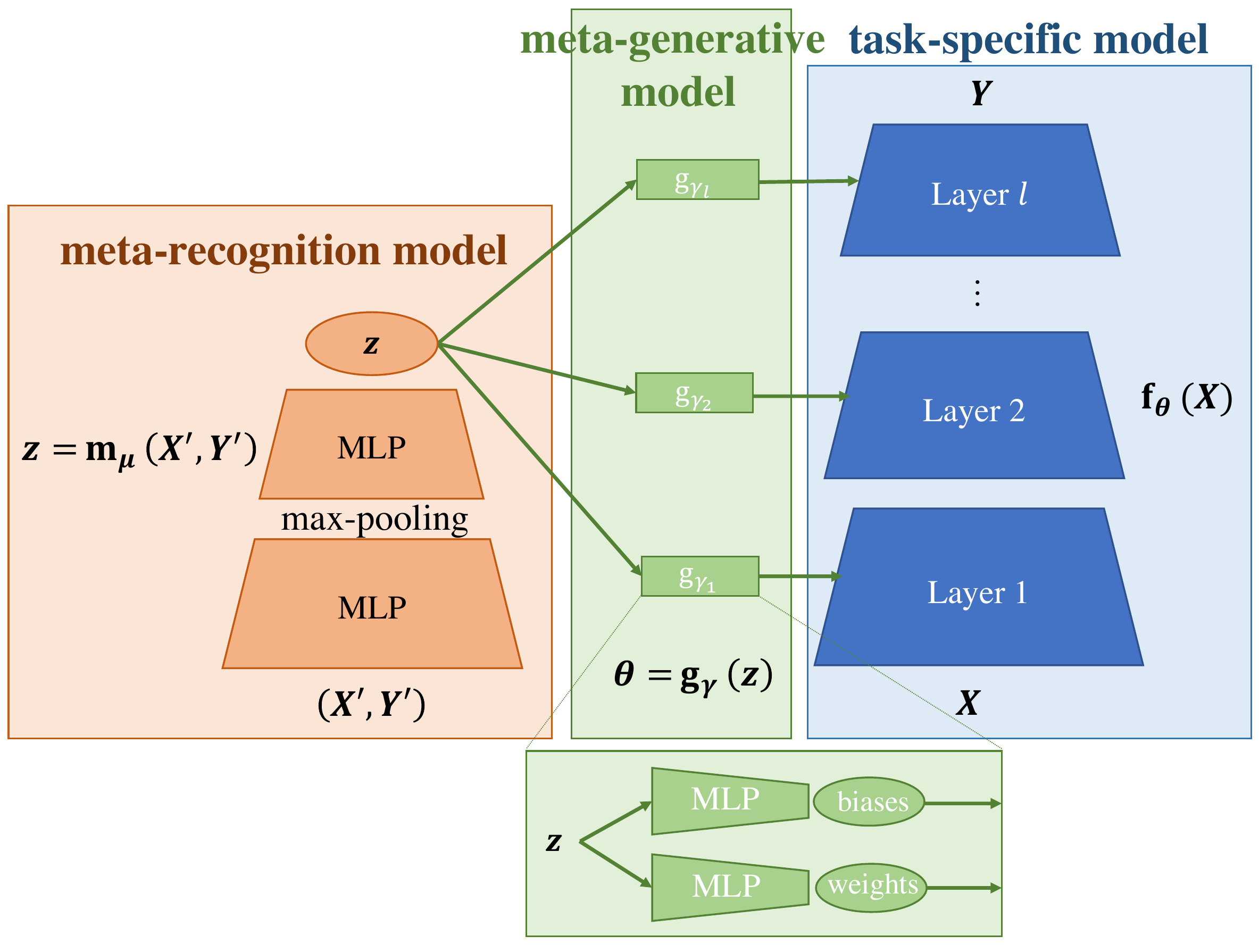}}
\caption{Architecture of our Meta-Learning Autoencoder (MeLA). MeLA augments a pre-existing neural network architecture $\f_{\phivec}$ (right) with a meta-recognition model (left) that generates the model code $\z$ based on a few examples $\X^{\prime}$, $\Y^{\prime}$, and a meta-generative model (middle) that generates the parameters $\phivec$ of model $\f_{\phivec}$ based on the model code. $\f_{\phivec}$, $\g_{\gammavec}$ and $\m_{\muvec}$ are implemented as multilayer perceptron (MLP).}
\label{ArchitectureFig}
\end{center}
\vskip -0.15in
\end{figure}

\paragraph{Meta-learning autoencoder architecture}
\label{sec:architecture}

The architecture of our Meta-Learning Autoencoder (MeLA) is illustrated in \fig{ArchitectureFig}.
It is defined by three vector-valued functions $\f_{\phivec}$, $\g_{\gammavec}$ and $\m_{\muvec}$ that are defined by feedforward neural networks parametrized by vectors $\phivec$, $\gammavec$ and $\muvec$, respectively.
In contrast to prior methods for learning to quickly adapt to different datasets \cite{pmlr-v70-finn17a} or using memory-augmented setup \cite{pmlr-v48-santoro16,NIPS2016_6385}, the MeLA takes full advantage of the prior that the datasets are generated by a hidden model class, where the functions $\h$ lie in a relatively low-dimensional submanifold of the space of all functions. Based on this prior, we use a meta-recognition model $\m_{\muvec}$ that maps a whole dataset $\D=(\X,\Y)$ to a model code vector $\z$, and a meta-generative model $\g_{\gammavec}$ that maps $\z$ to the parameters vector $\phivec$ of the network implementing the function $\f_{\phivec}$. In other words, $\phivec=\g_{\gammavec}(\z)$, and for a specific dataset $(\X,\Y)$, $\f_{\phivec}$ can be instantiated by 
\begin{equation}
{\f_{\phivec}}
={\f_{{\g_{\gammavec}}(\z)}}
=\f_{{\g_{\gammavec}}({\m_{\muvec}}(\X,\Y))}.
\end{equation}

% This means that, given a few examples $(\x_i,\y_i)$ from a new dataset arranged into matrices $\X$ and $\Y$,
% the function-specific model code $\z$ is directly calculated by $\z={\m_{\muvec}}(\X,\Y)$, which is then forwarded to the meta-generative model ${\g_{\gammavec}}(\z)$ to produce a specific model $f_{\g_{\gammavec}(\z)}$ for this dataset without training. Both the meta-recognition model $\m_{\muvec}$ and meta-generative model $\g_{\gammavec}$ are parametrized by multilayer feedforward neural networks (also known as multilayer perceptrons, MLPs), utilizing the expressive power of neural networks for learning inter-dataset variability and similarity.

% Here should put a figure displaying the architecture. And an equation showing how weights are generated.
This architecture is designed so that it can easily transform a neural network that is originally intended to learn from a single task into an architecture that can perform meta/few-shot learning on a number of tasks, combining the knowledge of individual task architectures with MeLA's meta-learning power. If the original single-task model is $\f_{\phivec}$, then without changing the architecture of $\f_{\phivec}$, we can simply attach a meta-recognition model $\m_{\muvec}$ and a meta-generative model $\g_{\gammavec}$ that generates the parameters of $\f_{\phivec}$, and train on an ensemble of tasks. 

\paragraph{Network architecture examples} 

Although the MeLA architecture described above can be implemented with any 
choices whatsoever for the three feedforward neural networks that define the functions 
$\f_{\phivec}$, $\g_{\gammavec}$ and $\m_{\muvec}$,
let us consider simple specific implementations to build intuition and get ready for the numerical experiments.

%Here we take MLP $\f_{\phivec}$ as an example, and elaborate the details and key considerations of MeLA (\fig{ArchitectureFig}). $\f_{\phivec}$ with 3D or 4D input tensors are easily generalizable.
Suppose we implement the main model $\f_{\phivec}$ as a network with two hidden layers with $s_1$ and $s_2$ neurons, respectively. Its input size is $s_0$ and its output size is $\sout$. 
The meta-recognition model $\m_{\muvec}$ takes as input $\X$ and $\Y$ concatenated horizontally into a single
$N\times(s_0+\sout)$ matrix, where $N$ is the number of training examples at hand. 
The feedforward neural network implementing $\m_{\muvec}$ has 
two parts: the first is a series of layers that collectively transform the $N\times(s_0+\sout)$ input matrix into an $N\times \spool$ matrix, where $\spool$ is the number of output neurons in this first block (we typically use 200 to 400 below). Then a max-pooling operation is applied over the $N$ examples, transforming 
this $N\times \spool$ matrix into a single vector of length $\spool$. 
The meta-recognition model $\m_{\muvec}$ is thus defined independently of the number of  training examples $N$.
As will be explained in the ``Influence identification" subsection below, max-pooling is key to MeLA, forcing the meta-recognition model to learn to capture key characteristics in a few representative examples. 
The second block of the $\m_{\muvec}$ network is a multilayer feedforward neural network, which takes as input the max-pooled vector, and 
transforms it into a 
$\slat$-dimensional model code $\z$ that parametrizes the functional relationship between $\x$ and $\y$.
% For the second block of the $\m_{\muvec}$ network, which takes the max-pooled vector as input, we explore two alternative implementations.
% In what we term {\it standard MeLA}, the second part is simply a multilayer feedforward neural network 
% transforming the $\spool$-dimensional max-pooled vector into the 
% $\slat$-dimensional vector $\z$ of model code variables that parametrize the functional relationship between $\x$ and $\y$.
% In what we term {\it variational MeLA},
% the second part consists of two feedforward networks that predict the 
% mean and log-standard deviation of a multivariate Gaussian from which $\z$ is sampled.

The meta-generative model $\g_{\gammavec}$ takes as input the model code $\z$, 
% (or a sample thereof for thevariational-MeLA)
and for each layer in the main model $\f_{\phivec}$, it has two separate neural networks that map $\z$ to all the weight and bias parameters of that layer. 
% IMHO, TOO OBVIOUS TO BE WORTH WRITING OUT:
%In our simple example, the first layer of $\f_{\phivec}$ has a 
%$s_0\times s_1$ weight matrix and a bias vector of length $s_1$,  so 
%$\g_{\gammavec}$
%has a corresponding neural network that maps the $\slat$-dimensional model code vector $\z$
%into the 
% $1\times (s_{input}\times s_1)$ matrix, reshaped to $(s_{input}\times s_1)$ which becomes the weights for the first layer of $\f_{\phivec}$, and another MLP that maps the $z$ to a $1\times s_1$ matrix which becomes the bias of the first layer of $\f_{\phivec}$. 
 % The same goes for other layers. 
We typically implement each of these subnetworks of $\g_{\gammavec}$ using 2-3 hidden layers with 60 neurons each. Compared with the original $\f_{\phivec}$, this implies only a linear increase in the number of parameters, independent of the number of tasks.
% {\bf (WHY?)}. 
%Further bounding the number of new parameters by using $\g_{\gammavec}$ to generate partial weights of $\f_{\phivec}$ will be left for future work. 

\begin{algorithm}[t]
   \caption{\textbf{Meta-Training for MeLA}}
   \label{alg:standard}
\begin{algorithmic}
   \STATE {\bfseries Require} datasets $\{\D^\alpha\}=\{(\X^\alpha, \Y^\alpha)\}$, $\alpha=1,2,...a$ 
   \STATE {\bfseries Require $n$}: number of meta-iterations
   \STATE {\bfseries Require $\beta$}: learning rate hyperparameter
   \STATE 1: Initialize random parameters for $\m_{\muvec}, \g_{\gammavec}$.
   \STATE 2: $i\gets 0$
   \STATE 3: \textbf{while} $i<n$:
   \STATE 4: \ \ \ \ \ \ $\{\D^{\alpha^\prime}\}\gets \text{permute}(\{\D^\alpha\})$   //Randomly permute the order of datasets.
   \STATE 5: \ \ \ \ \ \ \textbf{for} $\D_j$ \textbf{in} $\{\D^{\alpha^\prime}\}$ \textbf{do}
   \STATE 6: \ \ \ \ \ \ \ \ \ \ \ \ Split $\D_j=(\X_j,\Y_j)$ into training examples $(\X_j^{\rm train},\Y_j^{\rm train})$ and testing examples 
   \STATE \ \ \ \ \ \ \ \ \ \ \ \ \ \ \ \  $(\X_j^{\rm test},\Y_j^{\rm test})$
   \STATE 7: \ \ \ \ \ \ \ \ \ \ \ \ $\z\gets \m_{\muvec}(\X_j^{\rm train},\Y_j^{\rm train})$
   \STATE 8: \ \ \ \ \ \ \ \ \ \ \ \ $\phivec\gets \g_{\gammavec}(\z)$
   \STATE 9: \ \ \ \ \ \ \ \ \ \ \ \ Update $\muvec\gets\muvec-\beta\nabla_{\muvec}\ell\left[\f_{\g_{\gammavec}(\z)}(\X_j^{\rm test}),\Y_j^{\rm test}\right]$ 
   \STATE \ \ \ \ \ \ \ \ \ \ \ \ \ \ \ \ \ \ \ \ \ \ \ \ \ \ \ \ \  $\gammavec\gets\gammavec-\beta\nabla_{\gammavec}\ell\left[\f_{\g_{\gammavec}(\z)}(\X_j^{\rm test}),\Y_j^{\rm test}\right]$ 
   \STATE 10: \ \ \ \ \ \textbf{end for}
   \STATE 11: \ \ \ \ \ $i\gets i+1$
   \STATE 12: \textbf{end while} 
%   \STATE {\bfseries Output:} $\m_{\muvec}, \g_{\gammavec}$
\end{algorithmic}
\end{algorithm}
% \f_{{\g_{\gammavec}}({\m_{\muvec}}(\X,\Y))}

\paragraph{MeLA's meta-training and evaluation}
\label{sec:learning}

The extension from the training on a single-task $\f_{\phivec}$ to MeLA is straightforward. Suppose that the loss function for the single-task is $\ell(\yhat,\y)$, with expected risk
$R_{\ell,\D_k}(\f_{\phivec})\equiv\mathbb{E}_{(\X,\Y)\sim \D_k}\left[\ell(\f_{\phivec}(\X),\Y)\right]$. Then the meta-expected risk for MeLA is

\begin{equation}
\label{eqn:meta_risk}
    R_{\ell,p(\D)}(\m_{\muvec}, \g_{\gammavec})=\mathbb{E}_{\D_k\sim p(\D)}\left[\mathbb{E}_{(\X,\Y)\sim \D_k}\left[\ell(\f_{\g_{\gammavec}(\z)}(\X),\Y)\right]\right],
\end{equation}
where $p(\D)$ is the distribution for datasets $\{\D_k\}$ generated by the hidden model class $\h$. The goal of meta-training is to learn the parameters for the meta-recognition model $\m_{\muvec}$ and meta-generative model $\g_{\gammavec}$ such that $R_{\ell,p(\D)}(\m_{\muvec},\g_{\gammavec})$ is minimized:

\begin{equation}
\label{eqn:minimize_meta_risk}
(\muvec,\gammavec)=\argmin{(\muvec,\gammavec)}R_{\ell,p(D)}(\m_{\muvec}, \g_{\gammavec})
\end{equation}

Algorithm \ref{alg:standard} illustrates the step-by-step meta-training process for MeLA implementing an empirical meta-risk minimization for Eq. (\ref{eqn:minimize_meta_risk}). In each iteration, the training dataset ensemble is randomly permuted, from which each dataset is selected once for inner-loop task-specific training. Inside the task-specific training, the training examples for each dataset are used for calculating the model code $\z=\m_{\mu}(\X^{\rm train},\Y^{\rm train})$, after which the model parameter vector $\phivec=\g_{\gammavec}(\z)$ and the testing examples are used to calculate the task-specific testing loss $\ell(\f_{\g_{\gammavec}(\z)}(\X^{\rm test}),\Y^{\rm test})$, from which the gradients w.r.t.~$\muvec$ and $\gammavec$ are computed and used for one-step of gradient descent for the meta-recognition model and meta-generative model. Note that here the task-specific testing loss in the training datasets serves as the training loss in the meta-training.

During the evaluation of MeLA, we use the held-out datasets unseen during the meta-training. For each held-out dataset, we split it into training and testing examples. The training examples is fed to MeLA and a task-specific model is generated without any gradient descent. Then we evaluate the task-specific model on the testing examples in the held-out datasets. We also evaluate whether the task-specific model can further improve with a few more steps of gradient descent.
% The difference between the standard and variational MeLA is that the former has only one MLP after the max-pooling in the meta-recognition model, while the latter has two MLPs after max-pooling, calculating the mean and log-standard deviation of a multivariate Gaussian from which $\z$ is sampled, further encouraging the meta-recognition model to discover independent variations in the model class.

\paragraph{Influence identification}
\label{sec:influence}

The max-pooling over examples in the meta-recognition model $\m_{\muvec}$ is key to MeLA, and also provides a natural way to identify the influence of each example on the model $\f_{\phivec}$. 
Typically, some examples are more useful than others in in determining the model. 
For example, suppose that we try to learn a function $\f_{\phivec}$ defining on $\R^2$ that equals 1 inside a polygon and 0 outside, 
with different polygons corresponding to different models parametrized by $\phivec$.
Then data points near the polygon vertices carry far more information about $\phivec$ than do points in the deep interior,
and the max-pooling over the dimension of examples forces the meta-recognition model to recognize those most useful points, and based on them perform computation that returns a model code that determines the whole polygon. 
Recall that max-pooling compresses $N\times\spool$ numbers into merely $\spool$, which means that for each column 
of the $N\times\spool$ matrix, only one of the $N$ examples takes the maximum value and hence contributes to this feature.
We therefore define the \emph{influence} of an example as

\begin{equation}
\label{eqn:influence}
\text{Influence}=\frac{\text{Number of columns where the example is maximal}}{\spool}
\end{equation}

The influence of each example can be interpreted as a percentage, since it lies in $[0,1]$, and the influences sum to 1 for all the examples in the dataset fed to the meta-recognition model.

\paragraph{Interactive learning}
\label{sec:propose}

In some situations, measurements are hard or costly to obtain. 
It is then helpful if we can do better than merely acquiring random examples, and instead determine in advance at which data points $\x_i$ to collect measurements $\y_i$ to glean as much information as possible about the correct function $\f$.
Specifically, suppose that we want to predict $\hat{\y}^*=\f_{\phivec}(\x^*)$ as accurately as possible at a given input point $\x^*$ where we have no training data. If before making our prediction, we have the option to measure $\y$ at one of 
several candidate points $\x_1^{\prime},\x_2^{\prime}, ...\x_n^{\prime}$, then which point shall we choose?

The MeLA architecture provides a natural way to answer this question. 
% PROBABLY NOT NEEDED:
%MeLA can be viewed as consisting of two main components: the meta-learning related models $\m_{\muvec}$ and $\g_{\gammavec}$, and the main model $\f_{\phivec}$. The former is for generating the task-specific model, and the latter is for running the model for each individual task.
%In order to be more precise for the prediction $\y^*$ at $\x^*$, 
We can first use $\f_{\phivec}$ to calculate the current predictions for $\y_1^{\prime},\y_2^{\prime}, ...\y_n^{\prime}$ at $\x_1^{\prime},\x_2^{\prime}, ...\x_n^{\prime}$ based on current model generated by $\phivec=\g_{\gammavec}(\z)$ and $\z=\m_{\muvec}(\X^{\prime},\Y^{\prime})$, where $\X^{\prime}$ and $\Y^{\prime}$ are the examples that are already given. Then we can fix the meta-parameters $\muvec$ and $\gammavec$, and calculate the \emph{sensitivity matrix} of $\y^*$ w.r.t.~each current prediction $\y_i^\prime$:

%\begin{equation}
%\label{eqn:sensitivity}
%\frac{\partial\y^*}{\partial \y^\prime_i}=
%\frac{\partial \f_{\g_{\gammavec}(\z)}(\x^*)}{\partial\z}
%\frac{\partial\z}{\partial \y^\prime_i}
%\end{equation}

\begin{equation}
\label{eqn:sensitivity}
\frac{\partial\y^*}{\partial \y^\prime_i}=
\J
\frac{\partial\z}{\partial \y^\prime_i},
\quad\hbox{where}
\quad\J\equiv\frac{\partial \f_{\g_{\gammavec}(\z)}(\x^*)}{\partial\z}.
\end{equation}

We can select the candidate point whose sensitivity matrix has the largest determinant, \ie, the point for which the measured data
carries the most information about the answer $\y^*$ that we want:
\begin{equation}
    \y_i^\prime=\argmax{y_i^\prime}\left|\frac{\partial\y^*}{\partial \y^\prime_i}\right|
\end{equation}
If we model our uncertainty about $\y^*$ as a multivariate Gaussian distribution, then this criterion maximizes the entropy reduction, \ie, the number of bits of information learned about 
$\y^*$ from the new measurement.
Note that with $\gammavec$ fixed and for a given $\x^*$, the Jacobian matrix $\J$
%\begin{equation*}
%\J\equiv\frac{\partial %\f_{\g_{\gammavec}(\z)}(\x^*)}{\partial\z}
%\end{equation*}
is independent of the different candidate inquiry inputs $\x_1^{\prime},\x_2^{\prime}, ...\x_n^{\prime}$. 
This means that  we can simply select the  candidate point that has the largest ``projection" of $\left|\J\frac{\partial\z}{\partial \y^\prime_i}\right|$ onto $\J$, requiring in total only one forward and one backward pass for all the candidate examples to obtain the gradient. This factorization emerges naturally from MeLA's architecture.
\section{Related work}

MeLA addresses the problem of meta-learning \cite{thrun2012learning, schmidhuber1987evolutionary, 287172}, where an important subfield is to quickly adapt to new tasks with one-shot or few-shot examples. A recent innovative meta-learning method MAML \cite{pmlr-v70-finn17a} optimizes the parameters of the model so that it is easy to fine-tune to individual tasks in a few gradient steps. Another class of methods focuses on learning a learning rule or update functions \cite{schmidhuber1987evolutionary,bengio1992optimization,andrychowicz2016learning}, or learning an update function from a single good initialization \cite{li2017meta, ravi2016optimization}. Compared to these methods that only learn a \emph{single} good initialization point or how to update from a single initialization point, our method learns recognition and generative models that can quickly determine the model code for the model, and directly propose the appropriate neural network parameters tailored for each task without the need of fine-tuning.

Another interesting class of few-shot learning methods uses memory-augmented networks. \cite{NIPS2016_6385} proposes matching nets for one-shot classification, which generates the probability distribution for the test example based on the support set using attention mechanisms, essentially learning a ``similarity" metric between the test example and the support set. \cite{pmlr-v48-santoro16} utilizes a neural Turing machines for few-shot learning, and \cite{duan2016rl,wang2016learning} learn fast reinforcement learning agents with recurrent policies using memory-augmented nets. In contrast to memory-augmented approaches, our model learns to distill features from representative examples and produces a model code, based on which it directly generates the parameters of the main model. This eliminates the need to store the examples for the support set, and allows a continuous generation of models, which is especially suitable for generating a continuum of regression models. Other few-shot learning techniques include using Siamese structures \cite{koch2015siamese} and evolutionary methods \cite{mengistu2016evolvability}.

Autoencoders are typically used for representation learning in a single dataset, and have only recently been applied to multiple datasets. The recent neural statistician work  \cite{edwards2016towards} applies the variational autoencoder approach to the encoding and generation of datasets. Compared to their work, our MeLA differs in the following aspects. Firstly, the problem is different. While in neural statistician, each example in the dataset is an instance of a class, in MeLA, we are dealing with datasets whose examples are $(\x,\y)$ pairs, where we don't know \emph{a priori} where the input $\x$ will be in testing time. Therefore, direct autoencoding of datasets is not enough for prediction, especially for regression tasks. Therefore, instead of using autoencoding to generate the \emph{dataset}, our MeLA uses autoencoding to generate the \emph{model} that can generate the dataset given test inputs $\X$, which is a more compact way to express the relationship between $\X$ and $\Y$.

The idea of using an indirect encoding for the weights of another network originates from the neuroevolution algorithm of HyperNEAT \cite{stanley2009hypercubef}. Both the structure and weights are updated by evolutionary algorithms. \cite{fernando2016convolution} improves this method by making the weights differentiable, and Hypernetworks \cite{ha2016hypernetworks} further make both the network generator and network differentible. Our MeLA gets inspirations from these prior works, and also differs in several key aspects: while the above works focus on learning a single task, MeLA is endowed with a recognition model, and is designed for meta- and few-shot learning for unseen tasks.

% Learning physics: KVAE (\cite{NIPS2017_6951}), Visual deanimation (\cite{NIPS2017_6620}), Visual Interaction Networks (\cite{NIPS2017_7040})

\section{Experiments}
\label{Experiments}
Let us now test the core desiderata of MeLA: can it transform a model that is originally intended for single-task learning into one that can quickly adapt to new tasks with few examples without training, and continue to improve with a few gradient steps?
The baseline we compare with is a single network pretrained to fit to all tasks, which during testing is fine-tuned to each individual task through further training. MeLA has the same main network architecture $\f_{\phivec}$ as this baseline network, supplemented by the meta-recognition and meta-generative models trained via Algorithm \ref{alg:standard}. 
We also compare with the state-of-the-art meta-learning algorithm MAML \cite{pmlr-v70-finn17a}, with the same network architecture $\f_{\phivec}$.
%, and with an oracle network that ``cheats" by being given parameters of the true model $\h$ for each example. 
In addition, we explore the two other MeLA capabilities: influence identification and interactive learning.
% capabilities to quantify the influence of each example and propose the next example to obtain. 

For all experiments, the true model $\h$ and its parameters are hidden from all algorithms, except for an {\it oracle} model which ``cheats" by getting access to 
the true model parameters for each example, thus providing an  upper bound on performance. The performance of each algorithm is then evaluated on previously unseen test datasets. 
For all experiments, the Adam optimizer \cite{kingma2014adam} with default parameters is used for training and fine-tuning during evaluation. \footnote[1]{The code for MeLA and experiments will be open-sourced upon acceptance of the paper.}

\paragraph{Simple regression problem}

We first demonstrate the 3 capabilities of MeLA via the same  simple regression problem previously studied with MAML \cite{pmlr-v70-finn17a}, where
the hidden function class is
% $h(x)=c_3 \cdot \text{tanh}(c_1\cdot x + c_2)+c_4$,
% where the parameters $c_1\sim U[0.5,1.5]$, $c_2,c_4\sim U[-1,1]$, $c_3\sim U[1,2]$ 
$h(x)=c_1\sin(x + c_2)$,
and the parameters $c_1\sim U[0.1,5.0]$, $c_2\sim U[0,\pi]$
are randomly generated for each dataset. For each dataset, 10 input points $x_i$ are sampled from $U[-5,5]$ as training examples and another 10 are sampled as testing examples. 
100 such datasets are presented for the algorithms during training. The baseline model is a 3-layer network where each hidden layer has 40 neurons with leakyReLU activation. 
%Based on it, we compare MeLA with MAML, as well as the oracle model which also receives the true parameter of the underlying model. The pretrained and oracle models are two baselines setting up the lower and upper performance bound.

% \begin{figure}[t]
% \centering
% \begin{subfigure}{.45\textwidth}
%   \centering
%   \includegraphics[width=1.05\linewidth]{fig_sin_quicklearn.pdf}
%   \caption{}
%   \label{fig:sub1}
% \end{subfigure}%
% \begin{subfigure}{.55\textwidth}
%   \centering
%   \includegraphics[width=1.05\linewidth]{fig_influence.pdf}
%   \caption{}
%   \label{fig:sub2}
% \end{subfigure}
% \caption{(a) MSE vs. number of gradient steps for MeLA, MAML, fine-tuning and oracle on 20000 testing datasets. By just seeing the examples and without training, MeLA propose a model that has MSE of 0.208, while MAML and fine-tune model has a starting MSE of 3.050 and 3.074, respectively. After 10 gradient steps, MeLA's MSE goes down to 0.129, while MAML and fine-tune go down to 0.139 and 2.382, respectively. The oracle's MSE is at around 0.012. (b) Different models' prediction with 0 gradient steps for an example test dataset. The markers's size is proportional to the influence identified by MeLA. Also plotted is the MeLA's prediction given only the top 3 influential examples.}
% \label{fig:sin_quicklearn}
% \end{figure}

\begin{figure}[ht] 
  \centering
   \includegraphics[width=0.9\linewidth]{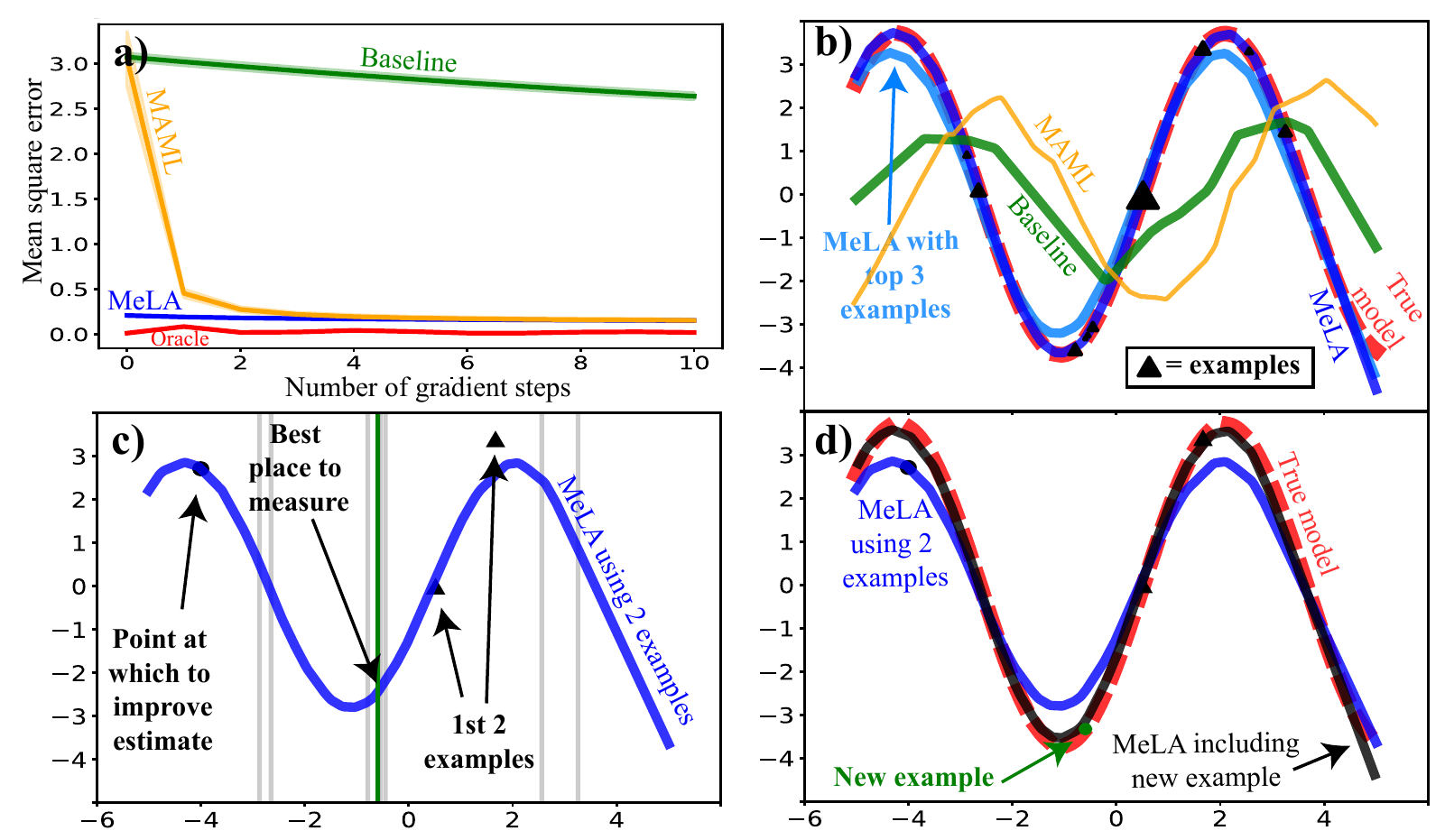} 
  \caption{(a) 
  MSE vs. number of gradient steps (with learning rate = 0.001, Adam optimizer) on 20,000 randomly sampled testing datasets, for MeLA, MAML, baseline (pretrained) and oracle. MeLA starts at MSE of 0.208 and gets down to 0.129 after 10  steps, while MAML starts at 3.05 and gets down to 0.208 after 5 steps. (b) Predictions after 0 gradient steps for an example test dataset (MAML is after 1 gradient step). The markers' size is proportional to the influence identified by MeLA. Also plotted is MeLA's prediction given only the top 3 influential examples. (c) To get a better prediction at $x^*=-4$ using only two examples at hand, MeLA requests the example at $x=-0.593$ from 8 candidate positions (vertical lines). (d) Improved estimate at $x^*=-4$ after obtaining the requested example. }
  \label{fig:sin} 
\end{figure}

The results are shown in Fig.~\ref{fig:sin}. Panel a) plots the mean squared error vs.~number of gradient steps on unseen randomly generated testing datasets, showing that MeLA outclasses the baseline model at all stages. 
It also shows that MeLA asymptotes to the same performance as MAML but learns much faster, starting with a low loss that MAML needs 5 gradient steps to surpass. 
Panel b) compares predictions with 0 gradient steps. MeLA not only proposes a model that accurately matches the true model, but also identifies each examples' influence on the model generation, and obtains good prediction if only the top 3 influential examples are given. Panels c) and d) show MeLA's capability of actively requesting informative examples by predicting which additional example will help improve the prediction the most. 
% In essence, {\bf move/drop next sentence to save space, since it's somewhat repeated above \& below??}sure 
% MeLA's meta-recognition model allows for quick recognition on informative examples to determine the model code, and quick generation of task-specific models via the meta-generative model.

\paragraph{Ball bouncing with state representation}
\label{sec:bounce-states}

\begin{figure}[t]
\centering
\begin{subfigure}{.48\textwidth}
  \centering
  \includegraphics[width=0.85\linewidth]{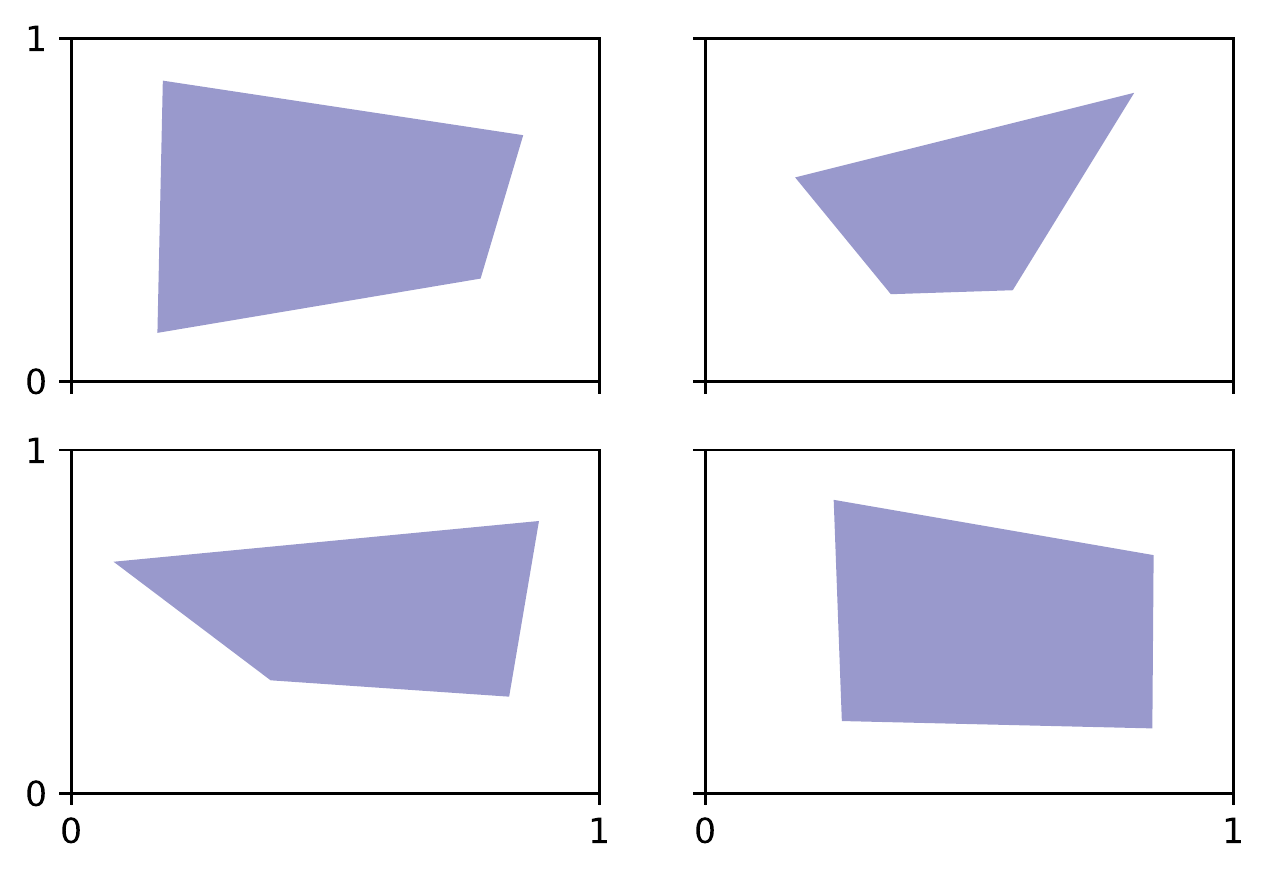}
  \caption{}
  \label{fig:bounce_visualize_1}
\end{subfigure}%
\begin{subfigure}{.52\textwidth}
  \centering
  \includegraphics[width=1.05\linewidth]{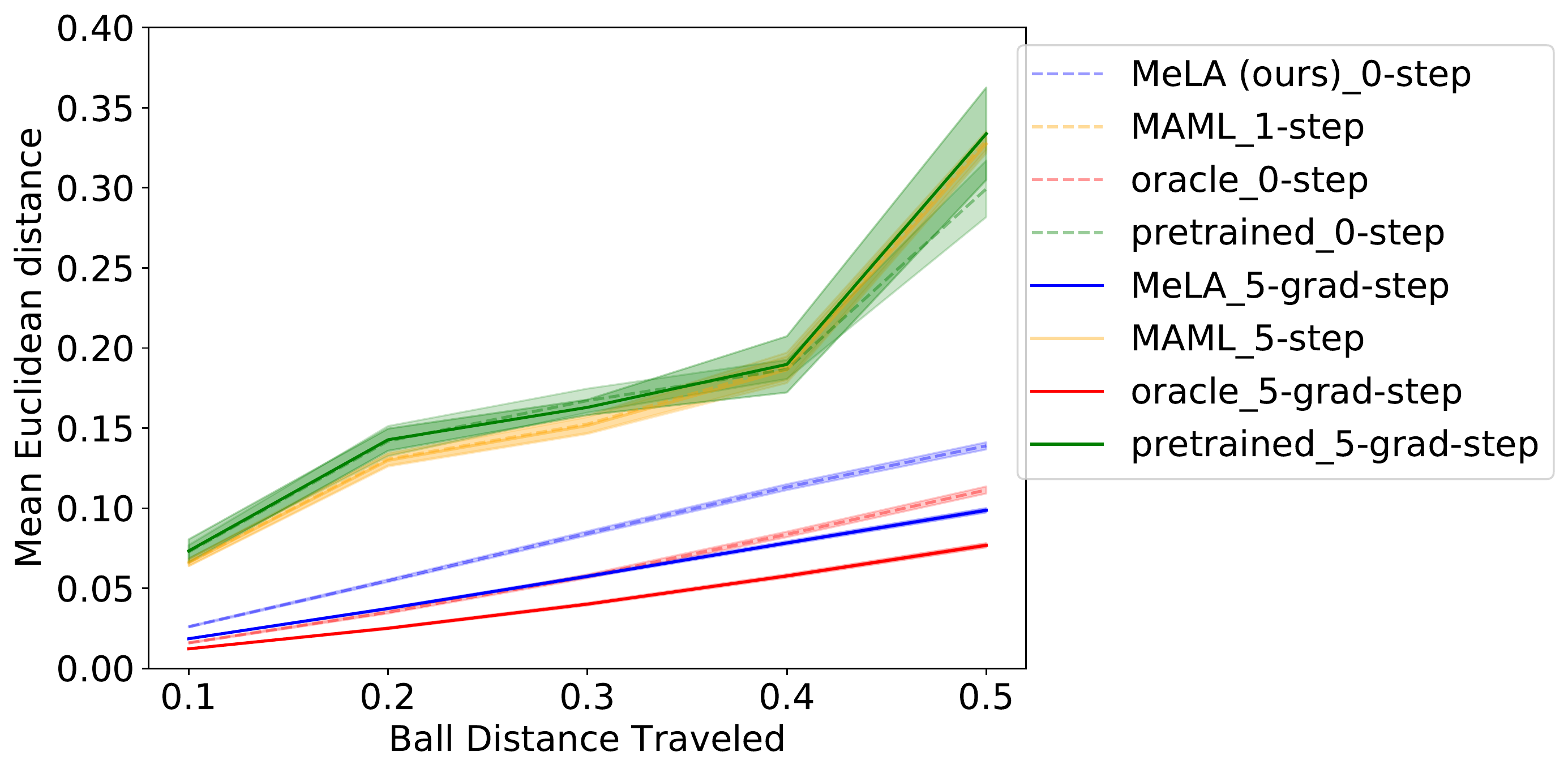}
  \caption{}
  \label{fig:bounce_visualize_2}
\end{subfigure}
\caption{(a) Examples of the polygon "bouncy-house" environments. (b) Mean Euclidean distance between target and prediction vs. rollout distance traveled 
%for baseline, MeLA, MAML {\bf (missing - WHAT ARE THE ODDS THAT JOHN WILL FINISH THIS ON TIME?)}, and oracle 
on 1000 randomly generated testing environments.}
\label{fig:bounce_visualize}
\end{figure}

Next, we test MeLA's capability in simple but challenging physical
environments, where it is desirable that an algorithm quickly adapts to each new environment with few observations of states or frames. 
Each environment consists of a room with 4 walls, whose frictionless floor is a random 4-sided convex polygon inside the  2-dimensional unit square $[0,1]\times[0,1]$ (Fig. \ref{fig:bounce_visualize}(a)), and a ball of radius 0.075
that bounces elastically off of these walls and otherwise moves with constant velocity. 
%The $x-$ and $y-$ coordinates of the ball are recorded every time it has moved a distance 0.1.
%
%Due to the large variability of the wall's positions and angles, 
Because the different room geometries give the ball conflicting bouncing dynamics in different environments, a model trained well in one environment may not necessarily perform well in another, providing an ideal test bed for meta- and few-shot learning. During training, all models take as input 3 consecutive time steps of ball's state 
 ($x-$ and $y-$ coordinates), recorded every time it has moved a distance 0.1.
The oracle model is also given as input the coordinates of the floor's 4 corners.

\begin{figure}[t]
\begin{center}
\vskip-3mm
\centerline{\includegraphics[width=0.6\columnwidth]{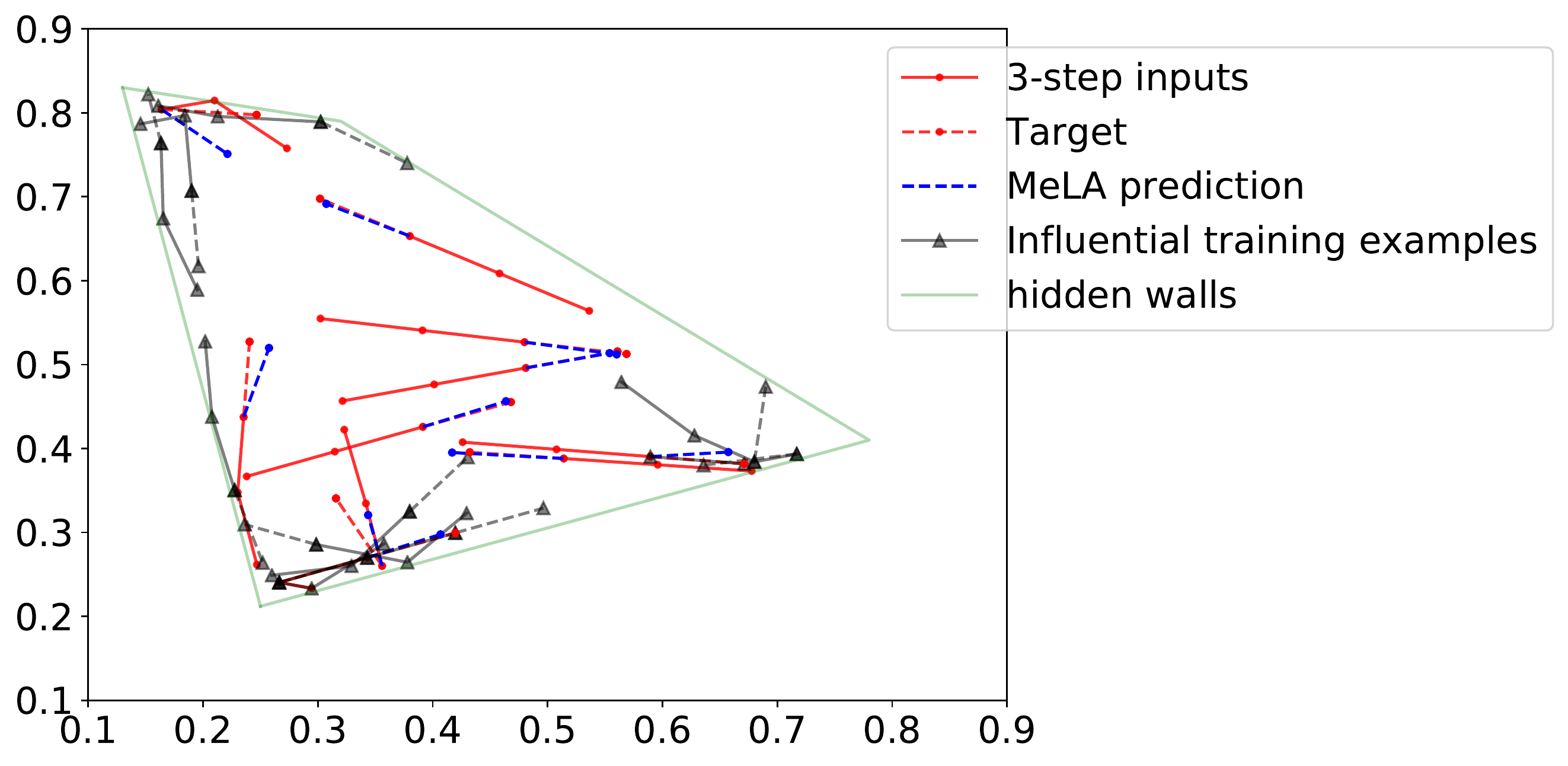}}
\vskip -0.05in
\caption{Ball bouncing prediction by MeLA for an example testing dataset. Also plotted are the top 10 most influential training  trajectories identified by MeLA, which are all near the vertices.}
\label{fig:bounce_influential}
\end{center}
\end{figure}

Fig. \ref{fig:bounce_visualize} (b) plots the mean Euclidean distance of the models' predictions vs. rollout distance traveled. We can see that MeLA outperforms pretrained and MAML for both 0 and 5 gradient steps. Moreover, what MeLA identifies as influential examples (Fig. \ref{fig:bounce_influential}) lies near the vertices of the polygon, showing that MeLA essentially learns to capture the convex hull of all the trajectories when proposing the model.

% Fig. 4: loss vs. gradient steps in testing tasks, for 50-shot learning. Comparing our model, MAML, and fine-tuning.

% Next, we explore MeLA 

% For video, representation, jointly training a convolutional autoencoder and the model.

% Fig. 4: loss vs. gradient steps in testing tasks, for 50-shot learning. Comparing our model, MAML, and fine-tuning.

% Fig. 5. Identifying the influence of each example.

% IN OUR REVISED VERSION THIS SUMMER, LETS ALSO PLOT THE BOUNDING POLYGON FOR THIS EXAMPLE! 
% Fig. 6. Proposing the next examples to obtain.

\paragraph{Video prediction}

To test MeLA's ability to integrate into other end-to-end architectures that deal with high-dimensional inputs, we present it with an ensemble of video prediction tasks, each of which has a ball bouncing inside randomly generated polygon walls. The environment setup is the same as in section \ref{sec:bounce-states}, except that the inputs are 3 consecutive frames of 39 x 39 pixel snapshots, and the target is a 39 x 39 snapshot of the next time step. For all the models, a convolutional autoencoder is used for autoencoding the frames, and the models differ only in the latent dynamics model that predicts the future latent variable based on the 3 steps of latent variables encoded by the autoencoder. For the pretrained model, a single 4-layer network with 40 neurons in each hidden layer is used for the latent dynamics model, training on all tasks. MAML and MeLA also have/generate the same architecture for the latent dynamics model. For the oracle model, the coordinates of the vertices are concatenated with the latent variables as inputs.

\begin{figure}[t]
\centering
\vskip -0.1in
\begin{subfigure}{.5\textwidth}
  \centering
  \includegraphics[width=0.98\linewidth]{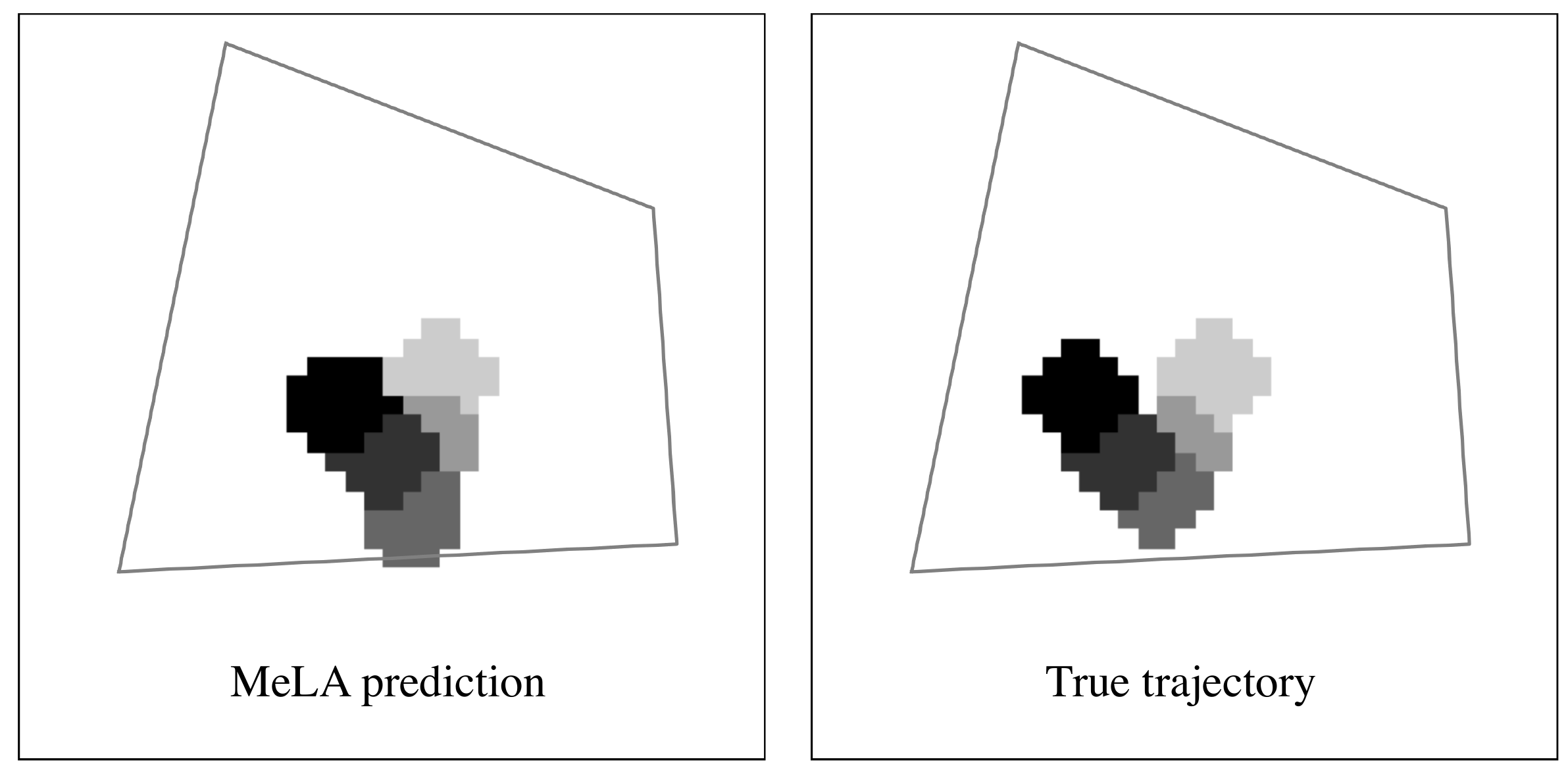}
  \caption{}
  \label{fig:sub1}
\end{subfigure}%
\begin{subfigure}{.5\textwidth}
  \centering
  \includegraphics[width=0.98\linewidth]{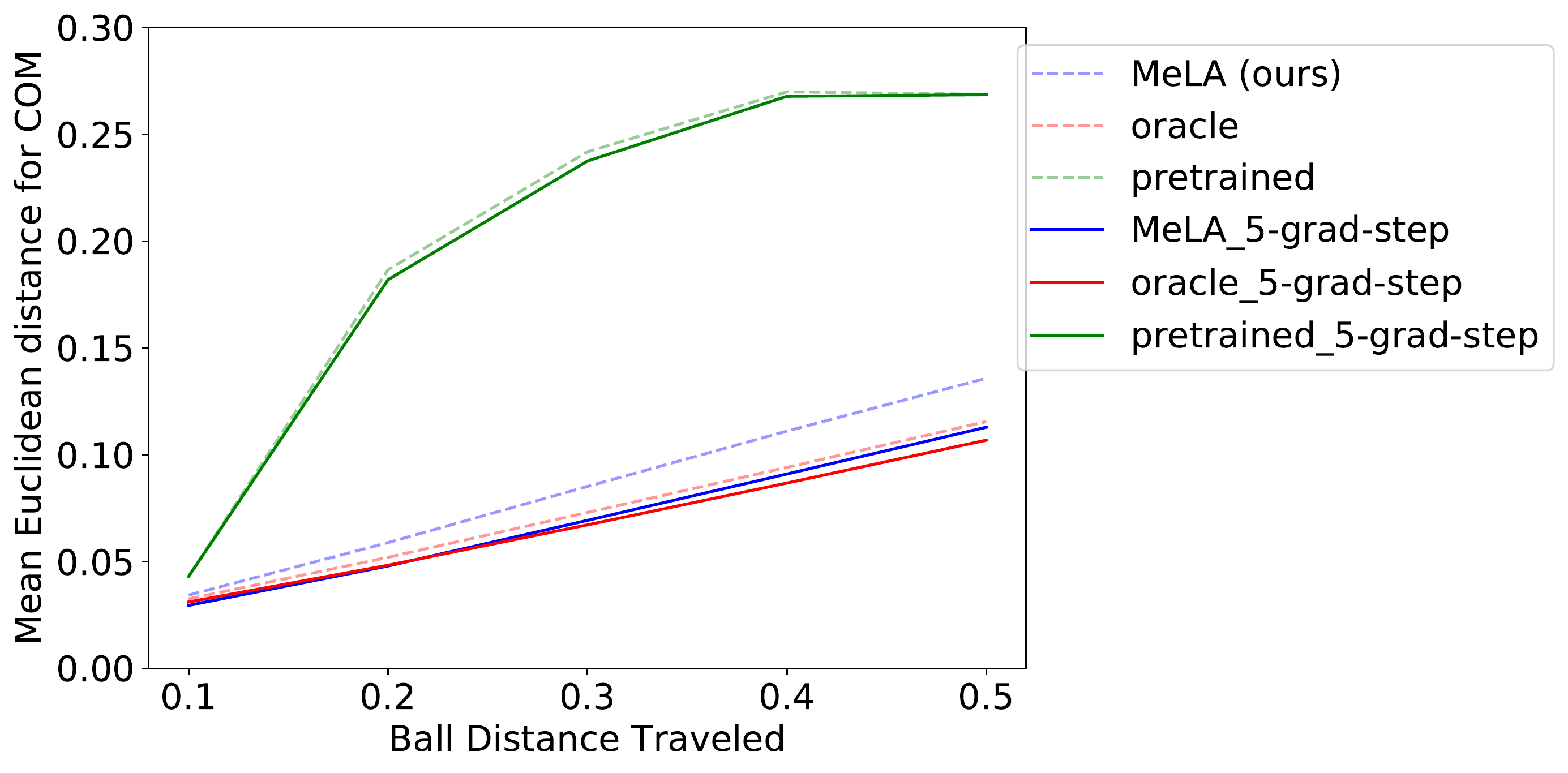}
  \caption{}
  \label{fig:bounce-images_perform}
\end{subfigure}
\caption{(a) Example MeLA prediction vs. true trajectory for 5 rollout steps, with an unseen testing environment without gradient steps. (b) Mean Euclidean distance between the center of mass (COM) of true trajectory and prediction vs. rollout distance traveled for MeLA, pretrained and oracle on 100 randomly generated testing environments.}
\label{fig:bounce_image_quicklearn}
\end{figure}

Fig.~\ref{fig:bounce-images_perform} plots the mean Euclidean distance of the center of mass of the models' predictions vs. rollout distance. We see that MeLA again greatly reduces the prediction error compared to the baseline model which has to use a single model to predict the trajectory in all environments. MeLA's accuracy is seen to be near that of the oracle, demonstrating that MeLA is learning to quickly recognize and model each environment and propose reasonable models.

\section{Conclusions}

In this paper, we have proposed MeLA, an algorithm for rapid recognition and determination of models in meta- and few-shot learning. We have demonstrated that MeLA can transform a model originally intended for single-task learning into one that can quickly adapt to new tasks with few examples, without training, and continue to improve with a few gradient steps. It learns better and faster than both the original model it is based on, and the state-of-the-art meta-learning algorithm MAML. We also demonstrate two additional capabilities of MeLA: its ability to identify influential examples, and how MeLA can interactively request informative examples to optimize learning.
% We have also demonstrated 
% MeLA's ability to preduct which additional data will be most valuable to acqquire to improve model prediction.

A core enabler of human ability to handle novel tasks is
our ability to quickly recognize and propose models in new environments, based on previously learned models. We believe that by incorporating this ability, machine learning models will become more adaptive and capable for new environments.
\newpage

\bibliography{reference}

\begin{thebibliography}{26}
\providecommand{\natexlab}[1]{#1}
\providecommand{\url}[1]{\texttt{#1}}
\expandafter\ifx\csname urlstyle\endcsname\relax
  \providecommand{\doi}[1]{doi: #1}\else
  \providecommand{\doi}{doi: \begingroup \urlstyle{rm}\Url}\fi

\bibitem[Andrychowicz et~al.(2016)Andrychowicz, Denil, Gomez, Hoffman, Pfau,
  Schaul, Shillingford, and De~Freitas]{andrychowicz2016learning}
Andrychowicz, M., Denil, M., Gomez, S., Hoffman, M.~W., Pfau, D., Schaul, T.,
  Shillingford, B., and De~Freitas, N.
\newblock Learning to learn by gradient descent by gradient descent.
\newblock In \emph{Advances in Neural Information Processing Systems}, pp.\
  3981--3989, 2016.

\bibitem[Bengio et~al.(1992)Bengio, Bengio, Cloutier, and
  Gecsei]{bengio1992optimization}
Bengio, S., Bengio, Y., Cloutier, J., and Gecsei, J.
\newblock On the optimization of a synaptic learning rule.
\newblock In \emph{Preprints Conf. Optimality in Artificial and Biological
  Neural Networks}, pp.\  6--8. Univ. of Texas, 1992.

\bibitem[Chang et~al.(2016)Chang, Ullman, Torralba, and
  Tenenbaum]{chang2016compositional}
Chang, M.~B., Ullman, T., Torralba, A., and Tenenbaum, J.~B.
\newblock A compositional object-based approach to learning physical dynamics.
\newblock \emph{arXiv preprint arXiv:1612.00341}, 2016.

\bibitem[Duan et~al.(2016)Duan, Schulman, Chen, Bartlett, Sutskever, and
  Abbeel]{duan2016rl}
Duan, Y., Schulman, J., Chen, X., Bartlett, P.~L., Sutskever, I., and Abbeel,
  P.
\newblock Rl2: Fast reinforcement learning via slow reinforcement learning.
\newblock \emph{arXiv preprint arXiv:1611.02779}, 2016.

\bibitem[Edwards \& Storkey(2016)Edwards and Storkey]{edwards2016towards}
Edwards, H. and Storkey, A.
\newblock Towards a neural statistician.
\newblock \emph{arXiv preprint arXiv:1606.02185}, 2016.

\bibitem[Fernando et~al.(2016)Fernando, Banarse, Reynolds, Besse, Pfau,
  Jaderberg, Lanctot, and Wierstra]{fernando2016convolution}
Fernando, C., Banarse, D., Reynolds, M., Besse, F., Pfau, D., Jaderberg, M.,
  Lanctot, M., and Wierstra, D.
\newblock Convolution by evolution: Differentiable pattern producing networks.
\newblock In \emph{Proceedings of the Genetic and Evolutionary Computation
  Conference 2016}, pp.\  109--116. ACM, 2016.

\bibitem[Finn et~al.(2017)Finn, Abbeel, and Levine]{pmlr-v70-finn17a}
Finn, C., Abbeel, P., and Levine, S.
\newblock Model-agnostic meta-learning for fast adaptation of deep networks.
\newblock In Precup, D. and Teh, Y.~W. (eds.), \emph{Proceedings of the 34th
  International Conference on Machine Learning}, volume~70 of \emph{Proceedings
  of Machine Learning Research}, pp.\  1126--1135, International Convention
  Centre, Sydney, Australia, 06--11 Aug 2017. PMLR.

\bibitem[Fraccaro et~al.(2017)Fraccaro, Kamronn, Paquet, and
  Winther]{NIPS2017_6951}
Fraccaro, M., Kamronn, S., Paquet, U., and Winther, O.
\newblock A disentangled recognition and nonlinear dynamics model for
  unsupervised learning.
\newblock In Guyon, I., Luxburg, U.~V., Bengio, S., Wallach, H., Fergus, R.,
  Vishwanathan, S., and Garnett, R. (eds.), \emph{Advances in Neural
  Information Processing Systems 30}, pp.\  3601--3610. Curran Associates,
  Inc., 2017.

\bibitem[Ha et~al.(2016)Ha, Dai, and Le]{ha2016hypernetworks}
Ha, D., Dai, A., and Le, Q.~V.
\newblock Hypernetworks.
\newblock \emph{arXiv preprint arXiv:1609.09106}, 2016.

\bibitem[Kansky et~al.(2017)Kansky, Silver, M{\'e}ly, Eldawy,
  L{\'a}zaro-Gredilla, Lou, Dorfman, Sidor, Phoenix, and
  George]{Kansky2017SchemaNZ}
Kansky, K., Silver, T., M{\'e}ly, D.~A., Eldawy, M., L{\'a}zaro-Gredilla, M.,
  Lou, X., Dorfman, N., Sidor, S., Phoenix, D.~S., and George, D.
\newblock Schema networks: Zero-shot transfer with a generative causal model of
  intuitive physics.
\newblock In \emph{ICML}, 2017.

\bibitem[Kingma \& Ba(2014)Kingma and Ba]{kingma2014adam}
Kingma, D.~P. and Ba, J.
\newblock Adam: A method for stochastic optimization.
\newblock \emph{arXiv preprint arXiv:1412.6980}, 2014.

\bibitem[Koch et~al.(2015)Koch, Zemel, and Salakhutdinov]{koch2015siamese}
Koch, G., Zemel, R., and Salakhutdinov, R.
\newblock Siamese neural networks for one-shot image recognition.
\newblock In \emph{ICML Deep Learning Workshop}, volume~2, 2015.

\bibitem[Li et~al.(2017)Li, Zhou, Chen, and Li]{li2017meta}
Li, Z., Zhou, F., Chen, F., and Li, H.
\newblock Meta-sgd: Learning to learn quickly for few shot learning.
\newblock \emph{arXiv preprint arXiv:1707.09835}, 2017.

\bibitem[Mengistu et~al.(2016)Mengistu, Lehman, and
  Clune]{mengistu2016evolvability}
Mengistu, H., Lehman, J., and Clune, J.
\newblock Evolvability search:directly selecting for evolvability in order to
  study and produce it.
\newblock In \emph{Proceedings of the Genetic and Evolutionary Computation
  Conference 2016}, pp.\  141--148. ACM, 2016.

\bibitem[Mnih et~al.(2015)Mnih, Kavukcuoglu, Silver, Rusu, Veness, Bellemare,
  Graves, Riedmiller, Fidjeland, Ostrovski, et~al.]{mnih2015human}
Mnih, V., Kavukcuoglu, K., Silver, D., Rusu, A.~A., Veness, J., Bellemare,
  M.~G., Graves, A., Riedmiller, M., Fidjeland, A.~K., Ostrovski, G., et~al.
\newblock Human-level control through deep reinforcement learning.
\newblock \emph{Nature}, 518\penalty0 (7540):\penalty0 529, 2015.

\bibitem[Naik \& Mammone(1992)Naik and Mammone]{287172}
Naik, D.~K. and Mammone, R.~J.
\newblock Meta-neural networks that learn by learning.
\newblock In \emph{[Proceedings 1992] IJCNN International Joint Conference on
  Neural Networks}, volume~1, pp.\  437--442 vol.1, Jun 1992.

\bibitem[Ravi \& Larochelle(2016)Ravi and Larochelle]{ravi2016optimization}
Ravi, S. and Larochelle, H.
\newblock Optimization as a model for few-shot learning.
\newblock 2016.

\bibitem[Santoro et~al.(2016)Santoro, Bartunov, Botvinick, Wierstra, and
  Lillicrap]{pmlr-v48-santoro16}
Santoro, A., Bartunov, S., Botvinick, M., Wierstra, D., and Lillicrap, T.
\newblock Meta-learning with memory-augmented neural networks.
\newblock In Balcan, M.~F. and Weinberger, K.~Q. (eds.), \emph{Proceedings of
  The 33rd International Conference on Machine Learning}, volume~48 of
  \emph{Proceedings of Machine Learning Research}, pp.\  1842--1850, New York,
  New York, USA, 20--22 Jun 2016. PMLR.

\bibitem[Schmidhuber(1987)]{schmidhuber1987evolutionary}
Schmidhuber, J.
\newblock \emph{Evolutionary principles in self-referential learning, or on
  learning how to learn: the meta-meta-... hook}.
\newblock PhD thesis, Technische Universit{\"a}t M{\"u}nchen, 1987.

\bibitem[Srivastava et~al.(2015)Srivastava, Mansimov, and
  Salakhutdinov]{Srivastava:2015:ULV:3045118.3045209}
Srivastava, N., Mansimov, E., and Salakhutdinov, R.
\newblock Unsupervised learning of video representations using lstms.
\newblock In \emph{Proceedings of the 32Nd International Conference on
  International Conference on Machine Learning - Volume 37}, ICML'15, pp.\
  843--852. JMLR.org, 2015.

\bibitem[Stanley et~al.(2009)Stanley, D'Ambrosio, and
  Gauci]{stanley2009hypercubef}
Stanley, K.~O., D'Ambrosio, D.~B., and Gauci, J.
\newblock A hypercube-based encoding for evolving large-scale neural networks.
\newblock \emph{Artificial life}, 15\penalty0 (2):\penalty0 185--212, 2009.

\bibitem[Thrun \& Pratt(2012)Thrun and Pratt]{thrun2012learning}
Thrun, S. and Pratt, L.
\newblock \emph{Learning to learn}.
\newblock Springer Science \& Business Media, 2012.

\bibitem[Vinyals et~al.(2016)Vinyals, Blundell, Lillicrap, kavukcuoglu, and
  Wierstra]{NIPS2016_6385}
Vinyals, O., Blundell, C., Lillicrap, T., kavukcuoglu, k., and Wierstra, D.
\newblock Matching networks for one shot learning.
\newblock In Lee, D.~D., Sugiyama, M., Luxburg, U.~V., Guyon, I., and Garnett,
  R. (eds.), \emph{Advances in Neural Information Processing Systems 29}, pp.\
  3630--3638. Curran Associates, Inc., 2016.

\bibitem[Wang et~al.(2016)Wang, Kurth-Nelson, Tirumala, Soyer, Leibo, Munos,
  Blundell, Kumaran, and Botvinick]{wang2016learning}
Wang, J.~X., Kurth-Nelson, Z., Tirumala, D., Soyer, H., Leibo, J.~Z., Munos,
  R., Blundell, C., Kumaran, D., and Botvinick, M.
\newblock Learning to reinforcement learn.
\newblock \emph{arXiv preprint arXiv:1611.05763}, 2016.

\bibitem[Watters et~al.(2017)Watters, Zoran, Weber, Battaglia, Pascanu, and
  Tacchetti]{NIPS2017_7040}
Watters, N., Zoran, D., Weber, T., Battaglia, P., Pascanu, R., and Tacchetti,
  A.
\newblock Visual interaction networks: Learning a physics simulator from video.
\newblock In Guyon, I., Luxburg, U.~V., Bengio, S., Wallach, H., Fergus, R.,
  Vishwanathan, S., and Garnett, R. (eds.), \emph{Advances in Neural
  Information Processing Systems 30}, pp.\  4539--4547. Curran Associates,
  Inc., 2017.

\bibitem[Wu et~al.(2017)Wu, Lu, Kohli, Freeman, and Tenenbaum]{NIPS2017_6620}
Wu, J., Lu, E., Kohli, P., Freeman, B., and Tenenbaum, J.
\newblock Learning to see physics via visual de-animation.
\newblock In Guyon, I., Luxburg, U.~V., Bengio, S., Wallach, H., Fergus, R.,
  Vishwanathan, S., and Garnett, R. (eds.), \emph{Advances in Neural
  Information Processing Systems 30}, pp.\  153--164. Curran Associates, Inc.,
  2017.

\end{thebibliography}
\bibliographystyle{icml2017}

\newpage
\appendix
\begin{center}
\begin{huge}
\textbf{Supplementals}
\end{huge}
\end{center}

\section{MeLA architectural details}

As described in section \ref{sec:architecture}, MeLA consists of a meta-recognition model $\m_{\muvec}$ and a meta-generative model $\g_{\gammavec}$ that generates the task-specific model $\f_{\phivec}$. The meta-recognition model consists of two blocks. The first block is a MLP with 3 hidden layers, each of which has 60 neurons with leakyReLU activation (unless otherwise specified, the leakyReLU activation in this paper all have a slope of 0.3 when the activation is below 0). The last layer has $\spool=200$ neurons and linear activation. Then a max-pooling is performed along the example dimension, collapsing the $N\times\spool$ matrix to $1\times\spool$ matrix, which feeds into the second block. The second block is an MLP with two hidden layers, each of which has 60 neurons with leakyReLU activation, and the last layer has $s_{\rm code}$ neurons with linear activation. The output is the model code $\z$. 

The meta-generative model $\g_{\gammavec}$ takes as input the model code $\z$, 
and for each layer in the main model $\f_{\phivec}$, it has two separate MLPs that map $\z$ to all the weight and bias parameters of that layer. For all the experiments in this paper, the MLPs in the meta-generative model have 3 hidden layers, each of which has 60 neurons with leakyReLU activation. The last layer of the MLP has linear activation, and has an output size equal to the size of weight or bias in the main network $\f_{\phivec}$. The output of each MLP in the meta-generative model is then reshaped into the size of the corresponding weight or bias matrix, and directly used as the parameters of $\f_{\phivec}$.

The architecture of the main network $\f_{\phivec}$ is dependent on the specific application, which MeLA's architecture is agnostic to. For the simple regression problem in this paper, we implement $\f_{\phivec}$ as an MLP with 2 hidden layers, each of which has 40 neurons with leakyReLU activation. The last layer has linear activation with output size of 1. For the ball bouncing with state representation experiment, $\f_{\phivec}$ is an MLP with input size of 6 and 3 hidden layers, each of which has 40 neurons with leakyReLU activation. The last layer has linear activation with output size of 2. For the video prediction task, the latent dynamics network uses the same architecture. The convolutional autoencoder used in this experiment is as follows. For the encoder, it has 3 convolutional layers with 32 $3\times 3$ kernels with stride 2 and leakyReLU activation. After that, it is flattened into 512 neurons, which feeds into a dense layer with 2 neurons and linear activation. For the decoder, the first layer is a dense layer with 512 neurons and linear activation, then the output is reshaped to a $N\times4\times4\times32$ tensor (32 is the number of channels). The tensor then goes into 3 layers of convolutional-transpose layers with 32 kernels, each with size of 3, stride of 2 and leakyReLU activation. For the leakyReLU activation in the convolutional autoencoder, we use a slope of 0.01 when the activation is below 0.

\end{document}